\pdfoutput=1

\documentclass[11pt]{article}

\usepackage[preprint]{acl}

\usepackage{times}
\usepackage{latexsym}

\usepackage[T1]{fontenc}

\usepackage[utf8]{inputenc}

\usepackage{microtype}

\usepackage{inconsolata}

\usepackage{graphicx}
\usepackage{amsmath} 
\title{Evaluating Expert Contributions in a MoE LLM for Quiz-Based Tasks}

\author{First Author \\
  Affiliation / Address line 1 \\
  Affiliation / Address line 2 \\
  Affiliation / Address line 3 \\
  \texttt{email@domain} \\\And
  Second Author \\
  Affiliation / Address line 1 \\
  Affiliation / Address line 2 \\
  Affiliation / Address line 3 \\
  \texttt{email@domain} \\}

\author{
 \textbf{Andrei Chernov \textsuperscript{1}},
\\
\\
 \textsuperscript{1}Independent Researcher,
\\
 \small{
   \textbf{Correspondence:} \href{mailto:chernov.andrey.998@gmail.com}{chernov.andrey.998@gmail.com}
 }
}

\begin{document}
\maketitle
\begin{abstract}
Recently, Large Language Models (LLMs) with Mixture of Experts (MoE) layers have gained significant attention. Currently, state-of-the-art LLMs utilize this architecture. There is a substantial amount of research on how to train such models and how to select hyperparameters for this architecture. However, there is a lack of studies focusing on post-evaluation analysis of MoE layer properties. In this paper, we take a first step toward closing this gap by evaluating expert contributions on the quiz-based MMLU benchmark. We show that most experts were never activated during inference on this benchmark. Additionally, the output distribution of gating networks is much closer to uniform than sparse. Finally, we demonstrate that the average performance of some experts within the same layer varies significantly.
\end{abstract}

\section{Introduction}

Recently, Large Language Models (LLMs) with Mixture of Experts (MoE) layers, instead of fully dense layers, have gained popularity \citep{du2022glam, wan2023efficient}. Currently, one of the best-performing models utilizes this architecture \citep{liu2024deepseek}. The main reason MoE models are preferred over dense models is that they tend to achieve similar performance while activating significantly fewer parameters, thereby reducing training time compared to dense LLMs \citep{muennighoff2024olmoe}. 

Most research on MoE in the natural language processing (NLP) domain has focused on either modifying the architecture to speed up inference—such as the Top-K gating mechanism \citep{shazeer2017outrageously}, which selects only the top-K experts with the highest probabilities—or adjusting the training loss to prevent the gating networks from always activating only a small subset of experts \citep{shazeer2017outrageously, shen2024jetmoe}.

In this paper, we focus on post-evaluation analysis of expert contributions to final predictions. Specifically, we evaluate the pretrained OLMoE model\footnote{\url{https://huggingface.co/allenai/OLMoE-1B-7B-0125-Instruct}} \citep{muennighoff2024olmoe} on the quiz-based MMLU benchmark \citep{hendrycks2020measuring} to address the following questions:

\begin{itemize}
    \item How many experts were activated at least once during inference on this benchmark?
    \item What does the distribution of gating network outputs look like? Does it tend to be sharp or closer to uniform?
    \item Do all experts perform equally in terms of accuracy?
\end{itemize}

\section{Experimental Setup}
In this paper, we investigate the contribution of each expert in the OLMoE model during inference on the MMLU benchmark. MMLU is a quiz-based benchmark that evaluates the knowledge and reasoning abilities of large language models (LLMs). It consists of $57$ datasets covering various domains, such as humanities, STEM, social sciences, and other fields.

We did not observe a significant difference in expert contributions across different domains. Therefore, in the results section (Section \ref{sec:results}), we present results aggregated over all datasets, comprising a total of $14,042$ questions.

For each question, the benchmark requires a model to select the correct answer from four possible choices: A, B, C, and D. Thus, the model needs to generate only one token corresponding to an answer. To assess the contribution of experts, we store the probabilities (alphas) from the gating network for each MoE layer when the model predicts the token corresponding to the correct answer.

The OLMoE model consists of $16$ MoE layers, each containing $64$ experts. For every question, we store an array of alphas with the following dimensions: $16 \times 64$. Note that only the top $8$ experts with the highest probabilities contribute to the final output.

To run the experiment, we utilized a V100 GPU with 16 GB of memory. We used a batch size of $2$, and the evaluation of the MMLU dataset took approximately $5$ hours.

\section{Results}

\label{sec:results}
\subsection{Distribution of Activated Experts}
\label{sec:res:dist}

In this section, we analyze how many experts were activated\footnote{To be activated, the corresponding probability for this expert from the gating function must be among the top $8$.} during inference, as well as the normalized distribution of activated experts for each datapoint. Tables \ref{tab:dist_1} and \ref{tab:dist_2} report the number of experts that were activated for at least one datapoint. 

Considering that the total number of experts is $64$, we observe that more than $60\%$ of the experts were never activated for the entire MMLU dataset. Additionally, we report the mean and standard deviation of natural entropy \citep{conrad2004probability}, defined as:

\begin{equation}
E = -\sum_{i \in \text{top } 8} p_i \log p_i
\end{equation}

where $p_i$ represents the normalized distribution over the highest gating probabilities, i.e., 

\begin{equation}
\label{eq:p_i}
p_i = \dfrac{\alpha_i}{\sum_{i \in \text{top } 8} \alpha_i},
\end{equation}
where $\alpha_i$ is the output from the gating network. We use natural entropy as a measure of uncertainty. It converges to zero when one expert has a probability close to $1$, meaning that only this expert contributes to the result. Conversely, when the distribution is uniform, entropy reaches its maximum value. Specifically, for a discrete distribution with $8$ outcomes, the highest entropy value is $2.0794$. 

Based on the reported entropy in the tables, we conclude that the distribution for each expert is far from sparse and instead tends to be closer to uniform. We believe this behavior is likely caused by auxiliary losses during the training procedure, which force the model to activate each expert approximately the same number of times. This prevents the model from converging to a small subset of preferred experts, thereby ensuring that all experts remain utilized. However, as our results suggest, this may lead to a gating probability distribution that is close to uniform, which might not be desirable. These results also hold for the distribution across all $64$ experts (see Appendix \ref{sec:appendix_a}).

A hypothesis that we believe is worth validating in future work is whether this uniform-like behavior negatively impacts the model's robustness. The primary concern is that the Top-K activation approach is not smooth. If the gating outputs follow a nearly uniform distribution, small changes in input may lead to significant differences in output due to a different set of experts being activated. Even if only the last expert in the top $K$ differs, this could still cause noticeable variations. As shown in the Table \ref{tab:top_8}, the weight of the eighth expert is significant, averaging $8.74\%$. This observation motivated us to investigate the average accuracy of each expert (see Section \ref{sec:res:acc}).

Additionally, an unexpected result for us is that entropy tends to increase from the first to the last layer. The first layer has the lowest entropy, while the last layer has one of the highest entropy. Intuitively, we expected the opposite: the last layer should be more confident in its predictions. One possible explanation is that some benchmark questions are too complex for the model, leading to less confident predictions. However, the standard deviation of entropy is low, indicating that the distribution remains stable across all questions, regardless of their complexity.


\begin{table*}
  \centering
  \resizebox{\textwidth}{!}{%
    \begin{tabular}{lcccccccc}
      \hline
      \textbf{Layer} & 1 & 2 & 3 & 4 & 5 & 6 & 7 & 8 \\
      \hline
      Mean Entropy (top 8) & $1.8516$ & $1.9375$ & $1.9297$ & $1.9531$ & $1.8516$ & $1.9219$ & $2.0156$ & $1.8984$ \\
      Std Entropy (top 8) & $0.0084$ & $0.0096$ & $0.0080$ & $0.0104$ & $0.0220$ & $0.0165$ & $0.0120$ & $0.0288$ \\
      Number of Activated Experts & $20$ & $14$ & $14$ & $10$ & $16$ & $14$ & $19$ & $15$ \\
      \hline
    \end{tabular}%
  }
  \caption{Statistical data per layer (Layers 1 to 8). Entropy calculated across the top 8 normalized experts.}
  \label{tab:dist_1}
\end{table*}

\begin{table*}
  \centering
  \resizebox{\textwidth}{!}{%
    \begin{tabular}{lcccccccc}
      \hline
      \textbf{Layer} & 9 & 10 & 11 & 12 & 13 & 14 & 15 & 16 \\
      \hline
      Mean Entropy (top 8) & $2.0000$ & $1.9297$ & $2.0469$ & $1.9688$ & $2.0156$ & $1.9063$ & $2.0156$ & $2.0313$ \\
      Std of Entropy (top 8) & $0.0092$ & $0.0245$ & $0.0092$ & $0.0233$ & $0.0135$ & $0.0354$ & $0.0151$ & $0.0133$ \\
      Number of Activated Experts & $19$ & $12$ & $11$ & $14$ & $15$ & $29$ & $24$ & $25$ \\
      \hline
    \end{tabular}%
  }
  \caption{Statistical data per layer (Layers 9 to 16). Entropy calculated across the top 8 normalized experts.}
  \label{tab:dist_2}
\end{table*}

\begin{table*}
  \centering
    \resizebox{\textwidth}{!}{%
  \begin{tabular}{lcccccccc}
    \hline
   \textbf{Top} & \textbf{1} & \textbf{2} & \textbf{3} & \textbf{4} & \textbf{5} & \textbf{6} & \textbf{7} & \textbf{8} \\
    \hline
    \textbf{Mean Probability} & $0.19297$ & $0.17575$ & $0.13789$ & $0.11344$ & $0.10380$ & $0.09713$ & $0.09181$ & $0.08721$ \\
    \textbf{Std Probability} & $0.01119$ & $0.01010$ & $0.01657$ & $0.00995$ & $0.00724$ & $0.00598$ & $0.00569$ & $0.00570$ \\
    \hline
  \end{tabular}
}
  \caption{Mean and standard deviation of top $8$ normalized probabilities from a gating network from the last MoE layer.}
  \label{tab:top_8}
\end{table*}

\subsection{Accuracy of each Expert}

\label{sec:res:acc}

In Section \ref{sec:res:dist}, we showed that the output distribution from the gating function is closer to uniform rather than sparse. This means that the contribution of each expert among the top $8$ is significant to the final outcome. In this section, we investigate whether all experts have similar accuracy or not. 

To achieve this, we compute the accuracy of each expert over all test data points where the expert was activated. Since an expert may contribute to different questions with varying weights, we also report the accuracy weighted by the probability assigned to each expert. Specifically, the weighted accuracy for expert $j$ is defined as:

\begin{equation}
\frac{\sum_{i=1}^{n} \alpha_{ij} \cdot \mathbf{1}(\hat{y}_i = y_i)}{\sum_{i=1}^{n} \alpha_{ij}},
\end{equation}

where $\alpha_{ij}$ represents the probability assigned to expert $j$ for datapoint $i$, and $\mathbf{1}(\hat{y}_i = y_i)$ is an indicator function that equals one when the final prediction is correct and zero otherwise. 

Additionally, we report the average contribution weight, computed as $100 \cdot p_i$ from Equation \ref{eq:p_i}, for each expert when it was activated. Results are presented for the first MoE layer (Table \ref{tab:1st_layer}) and the last MoE layer (Table \ref{tab:16th_layer}). In these tables, we include only experts that were activated in at least $1\%$ of the data (column: "Appearances"). There are $12$ such experts in the first MoE layer and $17$ in the last one.

For the first MoE layer, $7$ experts were activated in nearly all cases, meaning they appeared in more than $95\%$ of the data. The top eight experts were mainly chosen from three experts with indices:\footnote{The expert number refers to the index of an expert in an MoE layer, ranging from $0$ to $63$ inclusively.} $19$, $26$, and $52$. However, the accuracy of these experts varies significantly.

For the last MoE layer, only $3$ experts were activated in more than $95\%$ of the cases, providing the gating network with more flexibility in selecting different experts. In terms of accuracy, we observe a similar pattern to the first MoE layer: some experts achieve significantly higher accuracy than average (e.g., expert $12$), while others perform considerably worse (e.g., experts $34$ and $30$).

These findings suggest that a potential direction for future research could be adjusting the gating output probabilities by increasing the probability for high-accuracy experts and/or decreasing it for underperforming experts. This is particularly relevant given that the gating probability distribution is nearly uniform (see Section \ref{sec:res:dist}). This uniformity implies that the probability difference between high-accuracy experts and the top eight expert is relatively small. For instance, in the last MoE layer, the average gating function output for expert $12$, which performs significantly better than the average, is $0.0291$, while the average unnormalized probability for the top eight experts is $0.0317$.  

\begin{table*}
  \centering
  \resizebox{\textwidth}{!}{%
    \begin{tabular}{ccccc}
      \hline
      \textbf{Expert Number} & \textbf{Appearances (\%)} & \textbf{Accuracy (\%)} & \textbf{Weighted Accuracy (\%)} & \textbf{Mean Weight of the Expert (\%)} \\
      \hline
      $19$ & $30.93$ & $60.12$ & $60.17$ & $5.04$ \\
      $26$ & $38.38$ & $54.83$ & $54.72$ & $4.83$ \\
      $2$ & $99.70$ & $52.56$ & $52.52$ & $5.43$ \\
      $36$ & $99.82$ & $52.54$ & $52.48$ & $5.53$ \\
      $31$ & $100.00$ & $52.52$ & $52.49$ & $32.04$ \\
      $33$ & $100.00$ & $52.52$ & $52.56$ & $10.53$ \\
      $56$ & $100.00$ & $52.52$ & $52.49$ & $19.87$ \\
      $48$ & $100.00$ & $52.52$ & $52.60$ & $15.67$ \\
      $61$ & $98.72$ & $52.32$ & $52.48$ & $5.90$ \\
      $49$ & $2.20$ & $46.60$ & $46.44$ & $4.81$ \\
      $52$ & $26.36$ & $43.00$ & $43.35$ & $5.33$ \\
      $1$ & $1.69$ & $33.76$ & $33.63$ & $4.86$ \\
      \hline
    \end{tabular}%
  }
  \caption{Statistical data of experts in the first layer.}
  \label{tab:1st_layer}
\end{table*}

\begin{table*}
  \centering
  \resizebox{\textwidth}{!}{%
    \begin{tabular}{ccccc}
      \hline
      \textbf{Expert Number} & \textbf{Appearances (\%)} & \textbf{Accuracy (\%)} & \textbf{Weighted Accuracy (\%)} & \textbf{Mean Weight of the Expert (\%)} \\
      \hline
      $53$ & $1.44$ & $79.70$ & $80.16$ & $9.21$ \\
      $12$ & $27.30$ & $61.58$ & $62.56$ & $9.84$ \\
      $38$ & $10.68$ & $61.40$ & $62.17$ & $10.11$ \\
      $59$ & $5.60$ & $60.74$ & $62.23$ & $9.55$ \\
      $9$ & $46.00$ & $55.84$ & $55.82$ & $9.83$ \\
      $8$ & $81.80$ & $53.77$ & $55.44$ & $9.84$ \\
      $3$ & $85.50$ & $53.07$ & $54.09$ & $10.73$ \\
      $17$ & $100.00$ & $52.52$ & $51.45$ & $18.58$ \\
      $58$ & $100.00$ & $52.52$ & $52.49$ & $18.23$ \\
      $52$ & $99.00$ & $52.31$ & $51.23$ & $13.24$ \\
      $60$ & $73.10$ & $51.14$ & $51.24$ & $9.72$ \\
      $24$ & $50.64$ & $50.92$ & $51.73$ & $9.92$ \\
      $42$ & $1.17$ & $50.61$ & $51.04$ & $9.35$ \\
      $34$ & $62.38$ & $48.91$ & $48.98$ & $10.49$ \\
      $26$ & $5.07$ & $47.47$ & $47.98$ & $9.05$ \\
      $30$ & $47.51$ & $46.30$ & $46.62$ & $9.22$ \\
      $51$ & $1.18$ & $45.45$ & $47.13$ & $9.39$ \\
      \hline
    \end{tabular}%
  }
  \caption{Statistical data of experts in the 16th layer.}
  \label{tab:16th_layer}
\end{table*}

\section{Conclusion}
In this paper, we evaluated the contribution of experts in an LLM MoE model to the final output on a quiz-based benchmark. Our key findings are:

\begin{itemize}
    \item More than $60\%$ of experts were never activated during prediction. This implies that for quiz-based tasks, inactive experts can be removed, making the model smaller without any loss in performance. Additionally, this can significantly reduce training time during fine-tuning.
    
    \item The distribution of gating outputs is not sharp but rather nearly uniform across all MoE layers. Moreover, entropy does not decrease from the first layer to the last. Given that most LLM MoE models use a Top-K gating mechanism, which is a non-continuous gating method, this behavior may negatively impact the robustness of the models.
    
    \item Some experts perform better on average than others, suggesting that adjusting the gating output to prioritize high-accuracy experts could lead to performance improvements.
\end{itemize}

\newpage
\section*{Limitations}

The main limitation of this short paper is that the experiment was conducted on only one model and one benchmark. Our primary focus was on quiz-based datasets, and we believe that the MMLU benchmark represents this category well. Therefore, the use of a single benchmark is not a major limitation. However, a more significant limitation is that we evaluated only one LLM MoE model. We acknowledge that these results may not generalize to other LLM MoE models. 

The primary reason for using only one LLM MoE model is that most other models have a significantly larger number of parameters and require substantially more computational resources for inference, which we currently do not have.



\bibliography{main}

\newpage
\appendix

\section{Entropy of distribution across all Experts}
\label{sec:appendix_a}
In Table \ref{tab:all_entropy_1_8} and Table \ref{tab:all_entropy_9_16}, we show that all statements regarding entropy across the top $8$ experts in Section \ref{sec:res:dist} also hold for entropy across the probabilities of all $64$ experts given by the gating networks. Note that entropy generally increases with the number of possible outcomes, and for $64$ possible outcomes, the upper bound is $4.1589$.

\begin{table*}
  \centering
   \resizebox{\textwidth}{!}{%
  \begin{tabular}{lcccccccc}
    \hline
    & \textbf{1} & \textbf{2} & \textbf{3} & \textbf{4} & \textbf{5} & \textbf{6} & \textbf{7} & \textbf{8} \\
    \hline
    \textbf{Mean Entropy} & $3.78125$ & $3.89063$ & $3.85938$ & $3.85938$ & $3.75000$ & $3.75000$ & $3.87500$ & $3.75000$ \\
    \textbf{Std Entropy}  & $0.01245$ & $0.01447$ & $0.01056$ & $0.01453$ & $0.02759$ & $0.03113$ & $0.02478$ & $0.02783$ \\
    \hline
  \end{tabular}
  }
  \caption{Mean and standard deviation of entropy across all gating outputs (Layers 1 to 8).}
  \label{tab:all_entropy_1_8}
\end{table*}

\begin{table*}
  \centering
 \resizebox{\textwidth}{!}{%
  \begin{tabular}{lcccccccc}
    \hline
    & \textbf{9} & \textbf{10} & \textbf{11} & \textbf{12} & \textbf{13} & \textbf{14} & \textbf{15} & \textbf{16} \\
    \hline
    \textbf{Mean Entropy} & $3.79688$ & $3.51563$ & $3.62500$ & $3.46875$ & $3.50000$ & $3.51563$ & $3.64063$ & $3.82813$ \\
    \textbf{Std Entropy}  & $0.03467$ & $0.08057$ & $0.06152$ & $0.05884$ & $0.07666$ & $0.07422$ & $0.05835$ & $0.03809$ \\
    \hline
  \end{tabular}
  }
  \caption{Mean and standard deviation of entropy across all gating outputs (Layers 9 to 16).}
  \label{tab:all_entropy_9_16}
\end{table*}

\end{document}